\pdfoutput=1
\documentclass[11pt]{article}

\usepackage[]{EACL2023}
\usepackage{times}
\usepackage{latexsym}

\usepackage[T1]{fontenc}
\usepackage[utf8]{inputenc}

\usepackage{microtype}
\usepackage{graphicx}

\usepackage{inconsolata}
\usepackage{booktabs}
\usepackage{pgfplots}

\title{Large Language Models are few(1)-shot Table Reasoners}

\author{Wenhu Chen \\
  University of Waterloo, Vector Institute \\
  \texttt{wenhuchen@uwaterloo.ca} \\}

\begin{document}
\maketitle
\begin{abstract}
Recent literature has shown that large language models (LLMs) are generally excellent few-shot reasoners to solve text reasoning tasks. However, the capability of LLMs on table reasoning tasks is yet to be explored. In this paper, we aim at understanding how well LLMs can perform table-related tasks with few-shot in-context learning. Specifically, we evaluated LLMs on popular table QA and fact verification datasets like WikiTableQuestion, FetaQA, TabFact, and FEVEROUS and found that LLMs are competent at complex reasoning over table structures, though these models are not pre-trained on any table corpus. When combined with `chain of thoughts' prompting, LLMs can achieve very strong performance with only a 1-shot demonstration, even on par with some SoTA models. We show that LLMs are even more competent at generating comprehensive long-form answers on FetaQA than tuned T5-large. We further manually studied the reasoning chains elicited from LLMs and found that these reasoning chains are highly consistent with the underlying semantic form. We believe that LLMs can serve as a simple yet generic baseline for future research. The code and data are released in \url{https://github.com/wenhuchen/TableCoT}.
\end{abstract}

\section{Introduction}
The problem of structured knowledge grounding has been extensively studied for many years. Tables, as one of the most popular (semi)-structured forms to store world knowledge receive significant attention from the natural language processing (NLP) community. Traditional approaches mostly rely on synthesizing executable languages like SQL or SPARQL to access the information inside the table. However, these symbolic languages normally make a rigid assumption about the table and cannot capture the semantics of text chunks inside the table. Such issues are even more pronounced with web tables due to their irregular forms. To fully understand web tables, both structured reasoning and textual reasoning are required. Such challenges have attracted many researchers to work in the field. Recently, a wide range of table-based tasks have been proposed like table question answering~\cite{pasupat2015compositional,chen2020hybridqa,zhu2021tat,chen2021finqa,talmor2020multimodalqa,chen2020open,nan2022fetaqa}, table fact verification~\cite{chen2019tabfact,aly2021fact}, table-based generation~\cite{chen2020logical,parikh2020totto,nan2021dart}, and table-grounded conversation~\cite{budzianowski2018multiwoz,nakamura2022hybridialogue}. This wide range of table-based tasks all come with different input-output formats and domains. Due to the heterogeneity of these tasks, models achieving the best results on these tasks normally need to be fully fine-tuned on the specific downstream dataset with 10K-100K examples to achieve reasonable performance. 

Recently, there have been efforts like UnifiedSKG~\cite{UnifiedSKG} aiming to unify these heterogeneous table-based tasks as a generic text-to-text format. UnifiedSKG has shown that using T5-3B~\cite{raffel2020exploring} with the text-to-text format can already achieve state-of-the-art performance on almost all the table-based tasks without task-specific designs. However, the proposed text-to-text models still need to be fully fine-tuned on the downstream tasks. UnifiedSKG also identified that T0-style~\cite{sanh2022multitask} cross-task transfer can only achieve almost random performance.

\begin{figure}[t!]
  \centering
  \includegraphics[width=0.9\linewidth]{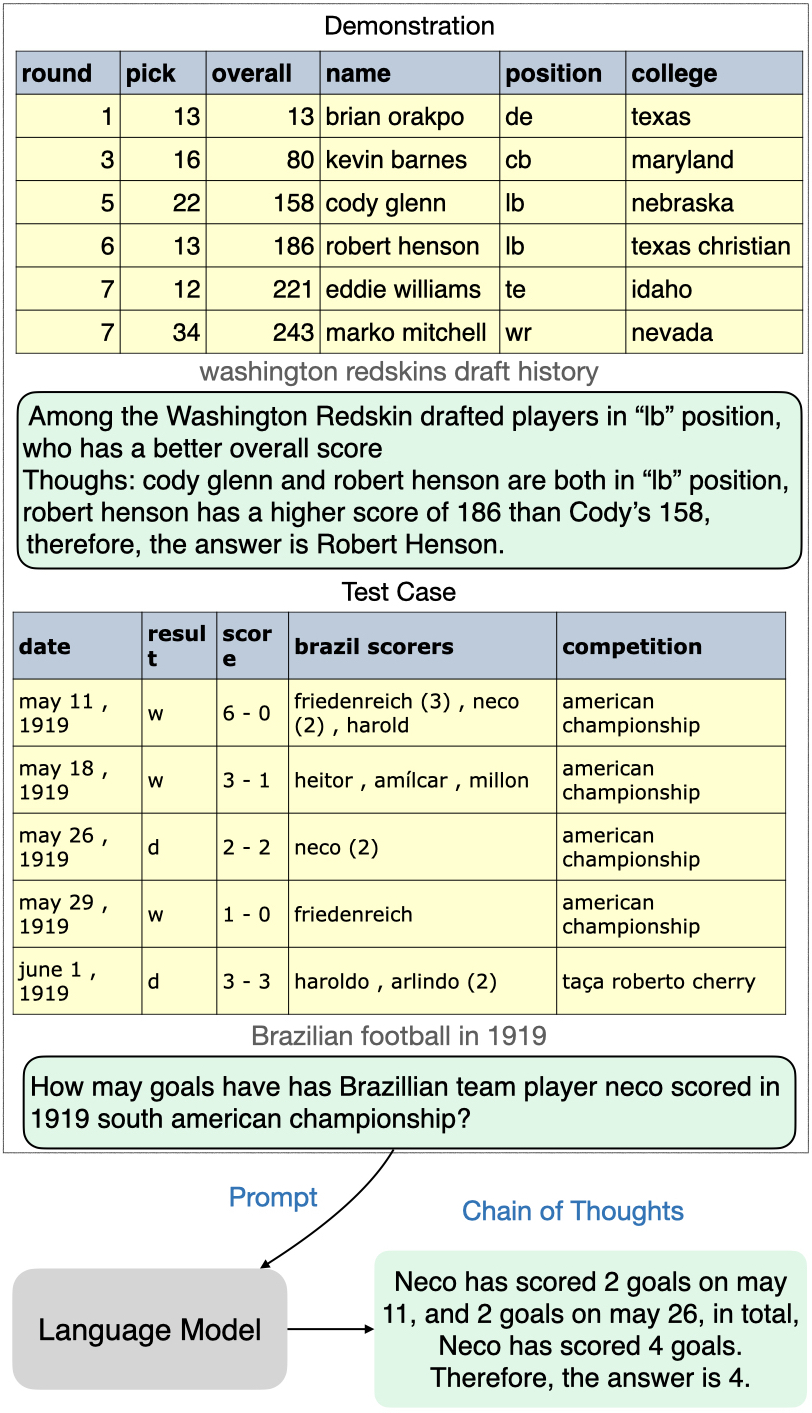}
  \caption{In-context learning for table-related tasks with chain-of-thoughts reasoning.}
  \label{fig:model}
\end{figure}

\citet{wei2022chain,wang2022self,zhou2022least,drozdov2022compositional} have recently discovered that large language models~\cite{brown2020language,chowdhery2022palm,ouyang2022training} can be used to solve complex mathematical and commonsense reasoning tasks with few-shot in-context learning. Inspired by this discovery, we aim at understanding whether these LLMs can also solve complex table-based reasoning tasks. Though the LLMs are not specifically designed to encode tables, given the enormous number of tables present in the pre-training corpus, we believe they are also competent at reasoning over table information.

In this paper, we experimented with few-shot in-context learning for LLMs as depicted in~\autoref{fig:model}. Instead of fine-tuning the model, we only provide a few examples to showcase the desired input-output format as the condition for the model to follow to solve unseen test examples. We experiment with several prompting variants including (1) direct prediction, (2) Chain of Thoughts~\cite{wei2022chain} (CoT), (3) Chains of thoughts with self-consistency~\cite{wang2022self} (CoT+SC). We evaluate these methods on WikiTableQA~\cite{pasupat2015compositional}, FetaQA~\cite{nan2022fetaqa}, TabFact~\cite{chen2019tabfact} and FEVEROUS~\cite{aly2021fact}. Our results reveal that LLMs~\cite{ouyang2022training,chen2021evaluating,chowdhery2022palm} can achieve striking performance with only 1 or 2 demonstrations, e.g. 48.8\% on WikiTableQuestions and 78.8\% on TabFact, which are on par some near-SoTA models~\cite{yu2021grappa,eisenschlos2020understanding}. On other datasets like FetaQA with long-form answers, our human evaluation reveals that GPT-3 can significantly outperform the fine-tuned T5-large by more than 30\% in terms of correctness and adequacy.

Furthermore, we manually studied the chain of thoughts elicited from LLMs and found that the rationale is highly consistent with the `ground truth' semantic forms when the model predictions are correct. We found that these models are surprisingly competent at performing symbolic operations over the table, like maximum, minimum, counting, comparison, addition, and difference. However, we also identify several issues of the LLMs on these table reasoning tasks: (1) due to the token limitation, the model is unable to generalize to `huge' tables with 30+ rows, which is the major error source, (2) LLMs can sometimes make simple mistakes when performing symbolic operations.

Due to the simplicity and generality, we believe LLMs with CoT should be used as an important baseline for any future table-related research.

\section{Related Work}
\subsection{Reasoning over Tables}
Table-based reasoning is traditionally accomplished by semantic parsing to execute commands on tables like WikiTableQuestions~\cite{pasupat2015compositional}, WikiSQL~\cite{zhong2017seq2sql}, and Spider~\cite{yu2018spider}. These models aim to synthesize SQL/SPARQL to interact with tables. However, these machine languages have a rigorous requirement regarding the tables, e.g. the value in the same column should follow the same data type. Such rigorous assumptions are frequently violated by web tables containing unnormalized free-form text in cells. Therefore, language understanding inside the table is essential to achieve a better score. Recently, ~\citet{yin2020tabert,herzig2020tapas,liu2021tapex,deng2022turl} have proposed to pre-train table and text to learn joint representation. These pre-trained models can use joint representation to perform reasoning implicitly without relying on symbolic execution. By pre-training the model on large-scale crawled or synthesized data, these models can normally achieve the best-known performance on table tasks. However, these models still require a significant amount of fine-tuning on the downstream datasets. Unlike these methods, we are interested in in-context learning, where the model can only learn with a few examples (demonstration) without any fine-tuning. One contemporary work similar to ours is BINDER~\cite{cheng2022binding}, which utilizes Codex to synthesize SQL to execute logical forms against tables for question answering. One big difference is that BINDER~\cite{cheng2022binding} involves logical form execution, if the execution fails, BINDER will fall back to using language models to answer the question, which is more similar to ours.

\subsection{In-context Learning with LLMs}
GPT-3~\cite{brown2020language} and other large language models demonstrated strong abilities to perform few-shot predictions without fine-tuning, where the model is given a description of the task in natural language with few examples. Scaling model size, data, and computing are crucial to enable this learning ability. Recently, ~\cite{rae2021scaling, smith2022using, chowdhery2022palm,du2022glam} have proposed to train different types of large language models with different training recipes. The LLMs have demonstrated a striking capability utilizing the few-shot prompts to accomplish unseen tasks without any fine-tuning, which is found to be an emergent capability not presented in smaller language models.  

\subsection{Chain of Thoughts Reasoning}
Although LLMs~\cite{brown2020language,chowdhery2022palm} have demonstrated remarkable success across a range of NLP tasks, their ability to demonstrate reasoning is often seen as a limitation. Such capability cannot be acquired simply by scaling up the model size. Recently, the `chain of thoughts' prompting~\cite{wei2022chain} has been discovered to empower LLMs to perform complex reasoning over text. By providing the model with several exemplars of reasoning chains, LLMs can learn to follow the template to solve difficult unseen tasks. Later, ~\citet{wang2022self} propose to use self-consistency with CoT to further improve performance. Later on, ~\citet{kojima2022large} discovered that LLMs can even perform reasoning without any demonstration by using appropriate prompts. These recent findings reveal the strong capability of LLMs to perform complex reasoning. However, the current studies are still heavily focused on text-based tasks like question answering, common sense reasoning, etc. The models' capability to reason over tables is yet unknown. In this paper, we are specifically interested in understanding LLMs' capability to reason over web tables with CoT prompting.

\begin{figure}[t!]
  \centering
  \includegraphics[width=1.0\linewidth]{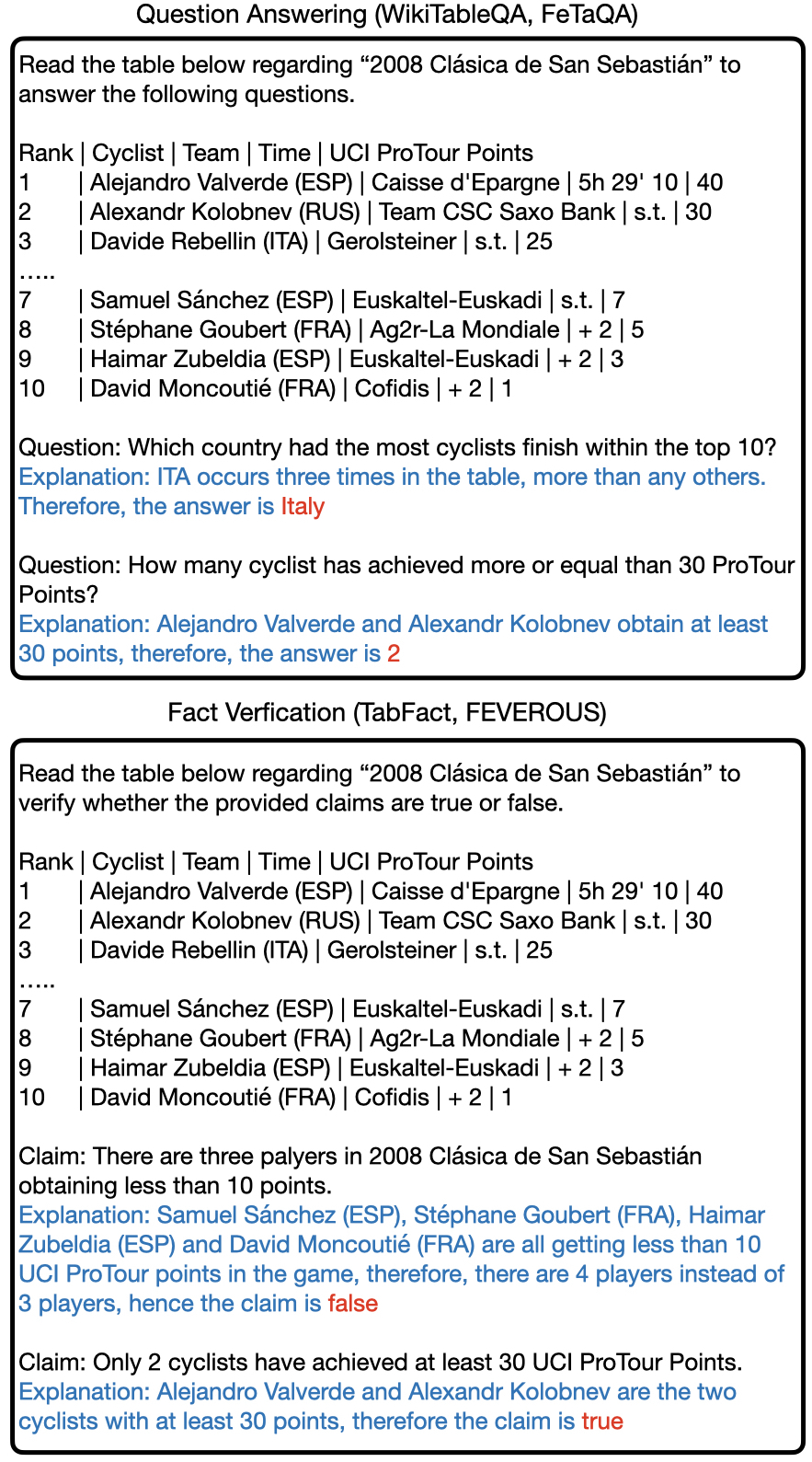}
  \caption{Prompts used for question answering and fact verification tasks.}
  \vspace{-2ex}
  \label{fig:prompt}
\end{figure}
\section{Method}
We experiment with different in-context learning methods to solve the table-based reasoning tasks. To formulate the prompt, we linearize the table and concatenate it with a few examples as demonstrations of the language model to predict the output from an unseen test example. The format is described in~\autoref{fig:prompt}. We mainly investigate three different variants for language model prompting, including (1) Direct Prediction, (2) Chain of Thoughts (CoT), and (3) Chain of Thoughts + Celf-Consistentcy decoding (CoT+SC). For self-consistency methods, we use LLMs to generate five diverse reasoning paths and then use majority voting to select the most voted answer.   

To limit the budget and constrain the input token length, we truncate the input tables to contain only the first 22 rows and the first 8 columns. For each cell, we truncate the word length to contain only the first 10 words. Through such truncation, we can restrict the input token length to within 2000 tokens. We will talk about the impact of input token length on the final performance.

\section{Experimental Results}
For the GPT-3 experiments, we used the four provided models, Ada, Babbage, Curie, and Davinci with 350M, 1.3B, 6.7B, and 175B parameters respectively. We mainly use Davinci-text-002~\cite{ouyang2022training} in our experiments. We also report results for Codex~\cite{chen2021evaluating} (Davinci-code-002) on some datasets. 
We use a temperature of 0.7 without any frequency penalty and without top-k truncation. We found that the model performance is robust to the sampling strategies and the hyper-parameters. These models are mainly trained on web-crawled data and code data, without any specialized training on table corpus.

\subsection{Datasets}
Here we list all of our datasets as follows:
\paragraph{WikiTableQuestions}
~\citet{pasupat2015compositional} consists of complex questions annotated based on Wikipedia tables. Crowd Workers are asked to compose a series of complex questions that include comparisons, superlatives, aggregation, or arithmetic operations. The annotated dataset is cross-validated by other crowd workers. In our experiments, we use the unseen test set for evaluation. We evaluate the standard test set with roughly 4000 questions. In this dataset, we adopt the answer exact match as our evaluation metric.

\paragraph{FetaQA}
~\citet{nan2022fetaqa} consists of free-form table questions. These questions are mostly complex questions that require integrating information from discontinuous chunks in the table. Instead of having short answers, the dataset annotates long free-form answers. Unlike other datasets using copies of short text spans from the source, the questions in FetaQA require a high-level understanding. We adopt sacre-BLEU and human evaluation as our evaluation metrics. The evaluation set contains a total of 2003 examples.

\paragraph{TabFact}
~\citet{chen2019tabfact} consists of both simple and complex claims annotated by crowd workers based on Wikipedia tables. In the simple subset, the claims normally do not involve higher-order operations like max/min/count, etc. While the complex subset mainly contains claims involving higher-order operations. We evaluate the original test set containing 12,779 examples. We report binary classification accuracy on the set.

\paragraph{FEVEROUS}
~\citet{aly2021fact} consists of compositional claims annotated by crowd workers regarding Wikipedia tables. Since the dataset contains both table-supported and text-supported claims. We filter out text-supported claims and only keep the 2,295 table-supported claims as our test set. Different from TabFact, FEVEROUS consists of more complex tables with irregular structures like multi-row, multi-column, multi-table, etc. We report dev-set accuracy. 

\subsection{Baselines}
In these experiments, we mainly consider the following baseline models.
\paragraph{Pre-trained Encoder-Decoder Model}
Pre-trained encoder-decoder model is one of our competitors, which aims to encode the table as a plain sequence into the encoder, and then apply the decoder to generate either an answer or a verdict. In this paper, we mainly compare against T5~\cite{raffel2020exploring} and BART~\cite{lewis2020bart} as our baselines.
\paragraph{Pre-trained Table Understanding Model}
This family of models is specifically pre-trained on the table-related corpus, which utilizes specific architecture to encode table structure and handle symbolic computation. In this paper, we mainly consider TAPAS~\cite{herzig2020tapas}, TABERT~\cite{yin2020tabert}, and TAPEX~\cite{liu2021tapex}. 
\paragraph{Neural Symbolic Model}
This family of models includes a non-pre-trained neural symbolic model, which can synthesize machine language to interact with the table. This line of work includes LogicFactChecker~\cite{zhong2020logicalfactchecker}, Neural-Symbolic Machine~\cite{liang2018memory}, etc.

\subsection{Main Results}
Here we show our main results for different datasets as follows.

\paragraph{WikiTableQuestions}
As can be seen from~~\autoref{tab:wikitq}, directly asking GPT-3 to generate answers can only lead to 26\% EM score. However, if we prompt the model with the CoT demonstrations, GPT-3 is more likely to follow the logical operation to derive the answers. With two demonstrations, GPT-3 can achieve roughly 46\% EM score. By switching from GPT-3 to Codex, we are able to further improve the EM score to over 48.8\%. These results are particularly surprising given that TAPAS has a built-in module to complete symbolic operations, while GPT-3 was not trained on any table-specific dataset. These results demonstrate GPT-3's built-in capabilities to perform diverse types of reasoning over tables. 
\begin{table}[!t]
    \small
    \centering
    \begin{tabular}{ccc}
    \toprule
    Type & Model &   Test EM \\
    \midrule
    Train & \citet{pasupat2015compositional}    &  37.1  \\
    Train & \citet{zhang2017macro}    &  43.7  \\
    Train & \citet{liang2018memory}  &  43.7  \\
    Train & \citet{agarwal2019learning}  &  44.1  \\
    Train & \citet{wang2019learning}  &  44.5  \\
    PT + FT & \citet{herzig2020tapas} & 48.8 \\
    PT + FT  & \citet{yu2021grappa} & \textbf{52.7} \\
    \midrule
    1-shot & GPT-3 Direct & 24.0  \\
    2-shot & GPT-3 Direct & 27.3  \\
    1-shot & GPT-3 CoT & 44.2  \\ 
    2-shot & GPT-3 CoT & 45.7  \\
    2-shot & Codex CoT & 48.8 \\
    \bottomrule
    \end{tabular}
    \caption{Experimental Results on WikiTableQuestions. PT means pre-training and FT means fine-tuning. }
    \label{tab:wikitq}
\end{table}

\paragraph{FetaQA}
As demonstrated in~\autoref{tab:fetaqa}, we compare GPT-3 with different fine-tuned models from~\citet{nan2022fetaqa}. Unlike the other datasets with short phrase answers, the goal of this dataset is to generate a complete long-form answer. Unlike WikiTableQuestion, the questions normally do not involve complex operations like max, min, compare, average, etc. The long-form answer is similar to the role of CoT. Therefore, we only applied `direct generation' in this experiment. In terms of BLEU score~\cite{papineni2002bleu}, GPT-3 is still a bit behind the fine-tuned T5-large.  However, the BLEU score cannot reflect the faithfulness and correctness of the model generation. Thus, we follow~\citet{nan2021dart} to do human evaluation over the four aspects: (1) fluency (whether the generated sentence contains the linguistic error), (2) correctness (whether the generated sentence answers the question correctly), (3) faithfulness (whether the generated sentence is grounded on the input table), and (4) adequacy (whether the generated sentence is comprehensive enough to cover all the answers). We list our results in~\autoref{tab:fetaqa_manual}. Similarly, we also sample 100 model predictions and manually evaluate their quality and adopt binary scores for each example.  As can be seen, GPT-3 can significantly outperform T5-large over all the aspects, i.e. more than 30\% improvement over correctness, adequacy, and faithfulness. The evaluation indicates that the model output is almost on par with the average human performance on this dataset. 

\begin{table}[!t]
    \small
    \centering
    \begin{tabular}{ccc}
    \toprule
    Type & Model  &   sacreBLEU \\
    \midrule
    zero-shot & Pipeline~\cite{nan2022fetaqa}    &  9.16  \\
    FT  & Pipeline~\cite{nan2022fetaqa}    &  11.00  \\
    FT  & T5-small~\cite{nan2022fetaqa} & 21.60 \\
    FT  & T5-base~\cite{nan2022fetaqa} & 28.14 \\
    FT  & T5-large~\cite{nan2022fetaqa} & \textbf{30.54} \\
    \midrule
    1-shot & GPT-3 Direct & 26.88  \\
    2-shot & GPT-3 Direct & \textbf{27.02}  \\
    \bottomrule
    \end{tabular}
    \caption{Experimental Results on FetaQA. PT means pre-training and FT means fine-tuning. }
    \label{tab:fetaqa}
\end{table}

\begin{table}[!t]
    \small
    \centering
    \begin{tabular}{ccccc}
    \toprule
    Source & Fluency  &   Correct &  Adequate & Faithful \\
    \midrule
    Pipeline &   85.2   & 25.4    &  23.6    &  23.6  \\
    T5-large       &  94.6    &  54.8   & 50.4   &  50.4  \\
    Human      &  95.0    & \textbf{92.4}    &  \textbf{95.6}  &  \textbf{95.6}  \\
    \midrule
    GPT-3     &  \textbf{98.0}  &  84.0   &  78.0  &  90.0  \\
    \bottomrule
    \end{tabular}
    \caption{Human Evaluation Results on FetaQA.}
    \label{tab:fetaqa_manual}
\end{table}

\paragraph{TabFact}
As demonstrated in~\autoref{tab:tabfact}, we compare GPT-3 against the other pre-trained and fine-tuned models including TAPAS~\cite{eisenschlos2020understanding}, TAPEX~\cite{liu2021tapex}, etc. We show that GPT-3 direct prediction is already getting a decent accuracy of 72\%, which is slightly higher than Logic FactChecker~\cite{zhong2020logicalfactchecker}. When combined with CoT reasoning, the model accuracy increases to over 77\%. Similar to before, we found that Codex can generate more accurate reasoning chains, thus achieving better accuracy of 78.8\%, which is only 2\% lower than pre-trained table understanding model TAPAS~\cite{eisenschlos2020understanding}. The more intriguing property about LLM + CoT is that the intermediate rationale can be produced without any training. All the existing trained models do not have the capability to produce the intermediate reasoning steps due to the lack of annotation in the dataset.

\begin{table}[!t]
    \small
    \centering
    \begin{tabular}{ccc}
    \toprule
    Type & Model    &   Overall\\
    \midrule
    FT  & \citet{chen2019tabfact}  &  65.1  \\
    FT  & \citet{zhong2020logicalfactchecker} & 71.1 \\
    FT  & \citet{zhang2020table} & 73.2 \\
    FT  &  \citet{yang2020program} & 74.4 \\
    FT  & \citet{lewis2020bart} &  82.5 \\
    PT + FT  & \citet{eisenschlos2020understanding} &  81.0 \\
    PT + FT  & \citet{liu2021tapex} & \textbf{84.2} \\
    \midrule
    1-shot & GPT-3 Direct  &  72.0            \\
    2-shot & GPT-3 Direct  &  73.9           \\
    1-shot & GPT-3 CoT     &  75.5           \\ 
    2-shot & GPT-3 CoT     &  76.0           \\
    1-shot & GPT-3 CoT+SC  &  77.3           \\
    2-shot & Codex CoT     &  78.8           \\
    \bottomrule
    \end{tabular}
    \caption{Experimental Results on TabFact. PT means pre-training and FT means fine-tuning. }
    \label{tab:tabfact}
\end{table}

\paragraph{FEVEROUS}
We demonstrate our results on FEVEROUS dev-set in~\autoref{tab:feverous} and compare different-sized UnifiedSKG models (built with T5). We found that GPT-3's performance with direct prediction is similar to UnifiedSKG-base. Similar to TabFact, we found that the model performance can be boosted with `chain of thoughts' prompting. The best-performing model is roughly between UnifiedSKG-base and UnifiedSKG-large. Compared to TabFact, the model's overall performance is weaker mainly because the table structure in FEVEROUS is more irregular, containing lots of segments and subtables. Such structural difficulties pose great challenges to GPT-3.
\begin{table}[!t]
    \small
    \centering
    \begin{tabular}{ccc}
    \toprule
    Type & Model    &   Dev Set \\
    \midrule
    FT  & \citet{aly2021fact}  &  82.23  \\
    FT  & UnifiedSKG-base~\cite{UnifiedSKG} &  75.05 \\
    FT  & UnifiedSKG-large~\cite{UnifiedSKG} & 79.81  \\
    FT  & UnifiedSKG-3B~\cite{UnifiedSKG} &  \textbf{82.40} \\
    \midrule
    1-shot & GPT-3 Direct  &  74.20  \\
    2-shot & GPT-3 Direct &  75.22  \\
    1-shot & GPT-3 CoT &  75.70 \\ 
    2-shot & GPT-3 CoT &  76.44  \\
    1-shot & GPT-3 CoT+SC &  \textbf{77.22}  \\
    \bottomrule
    \end{tabular}
    \caption{Experimental Results on FEVEROUS. PT means pre-training and FT means fine-tuning. }
    \vspace{-2ex}
    \label{tab:feverous}
\end{table}

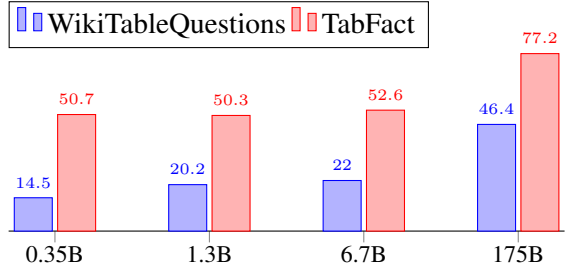
\begin{figure}[!t]
\begin{tikzpicture}
\begin{axis} [
ybar,
height=1.2in, 
bar width=0.5cm,
width=0.95\linewidth,
scale only axis,
ymin = 0, 
ymax = 100,
yticklabels=\empty,
axis x line*=bottom,
hide y axis,
xticklabel style = {font=\small,yshift=0.5ex},
symbolic x coords={
0.35B,
1.3B,
6.7B,
175B
},
legend style={
    at={(0,1.0)},
    anchor=north west,
    legend columns=-1
},
%title=Breakdown Accuracy over Question Types,
xtick=data,
yticklabels=\empty,
nodes near coords,
nodes near coords align={vertical},
every node near coord/.append style={font=\tiny},
]
\addplot coordinates {
(0.35B, 14.5) (1.3B, 20.2) (6.7B, 22.0) (175B, 46.4)
};
\addplot coordinates {
(0.35B, 50.7) (1.3B, 50.3) (6.7B, 52.6) (175B, 77.2)
};
\legend{WikiTableQuestions, TabFact}
\end{axis}
\end{tikzpicture}
\vspace{-4ex}
\caption{The model performance with respect to model size on WikiTableQuestions and TabFact.}
\vspace{-1ex}
\label{fig:scaling_law}
\end{figure}

\paragraph{Model Scaling}
We investigate the model scaling's impact on the final performance and plot our findings in~\autoref{fig:scaling_law}. On the WebTableQuestions dataset, we found that model size is essential for achieving the best performance. As can be seen, the 6.7B GPT-3 model is only achieving half of the performance of the 175B GPT-3 model. Similarly, on TabFact, we found that the smaller models with 6.7B or fewer parameters are almost getting random accuracy, which is even worse than QA tasks. This again suggests that LLMs' reasoning ability over web tables is emergent as the model scales up.

\subsection{Case Study}
\begin{figure}[t!]
  \centering
  \includegraphics[width=1.0\linewidth]{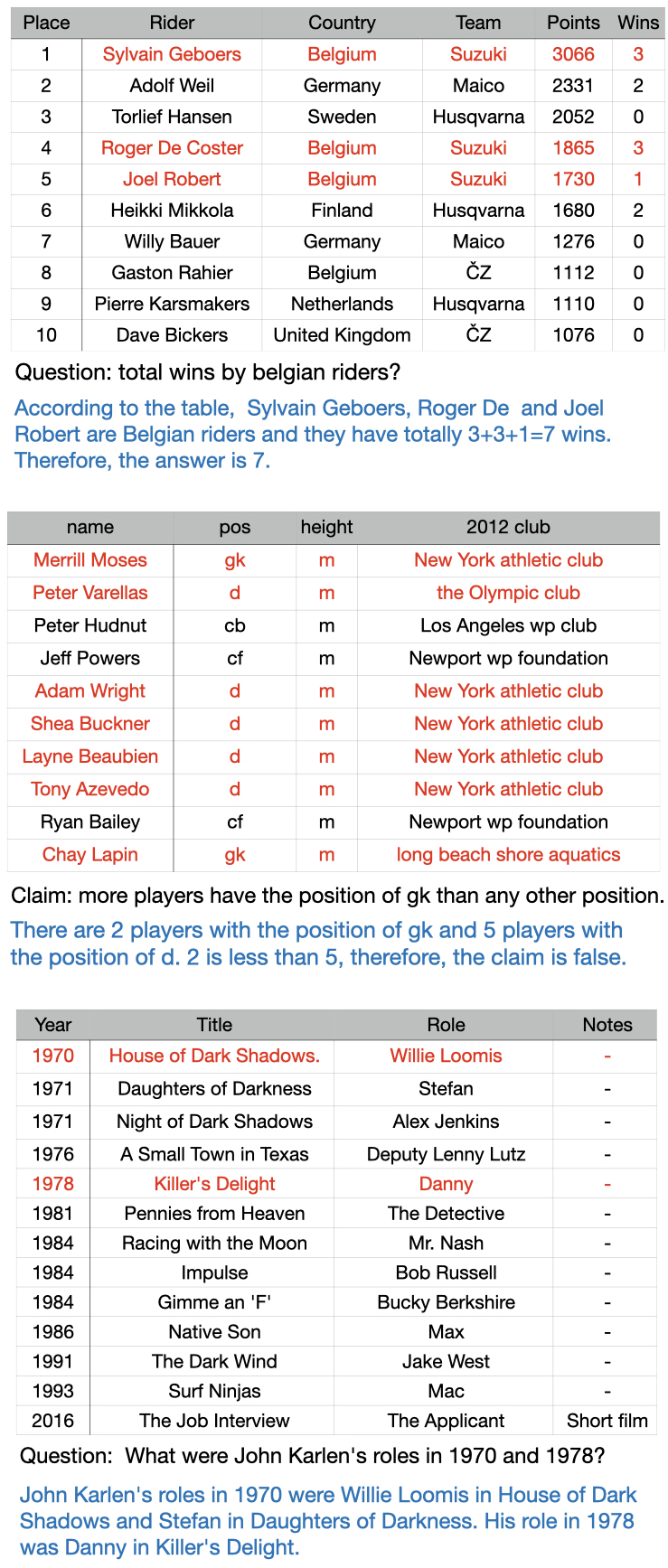}
  \caption{`Correct' predictions from WikiTableQuestions, TabFact, and FetaQA datasets, where the `blue' text are the outputs from the GPT-3, `red' means the correct rows to reference.}
  \label{fig:examples}
  \vspace{-2ex}
\end{figure}

We demonstrate a few examples in~\autoref{fig:examples} where GPT-3 makes correct predictions. In the first example, GPT-3 is able to first identify all the Belgian riders from the table and then perform the addition of 3+3+1=7 precisely. In the second example, GPT-3 can identify the players with the position of `d' and count the number correctly to refute a false claim. In the third example, we can see that GPT-3 is able to associate multiple blocks of information to generate a comprehensive long-form answer. The elicited `chain of thoughts' in these examples are highly aligned with the underlying semantic forms. These findings suggest that LLMs like GPT-3 can provide high-quality explanations to justify their decision-making.

\begin{figure}[t!]
  \centering
  \includegraphics[width=1.0\linewidth]{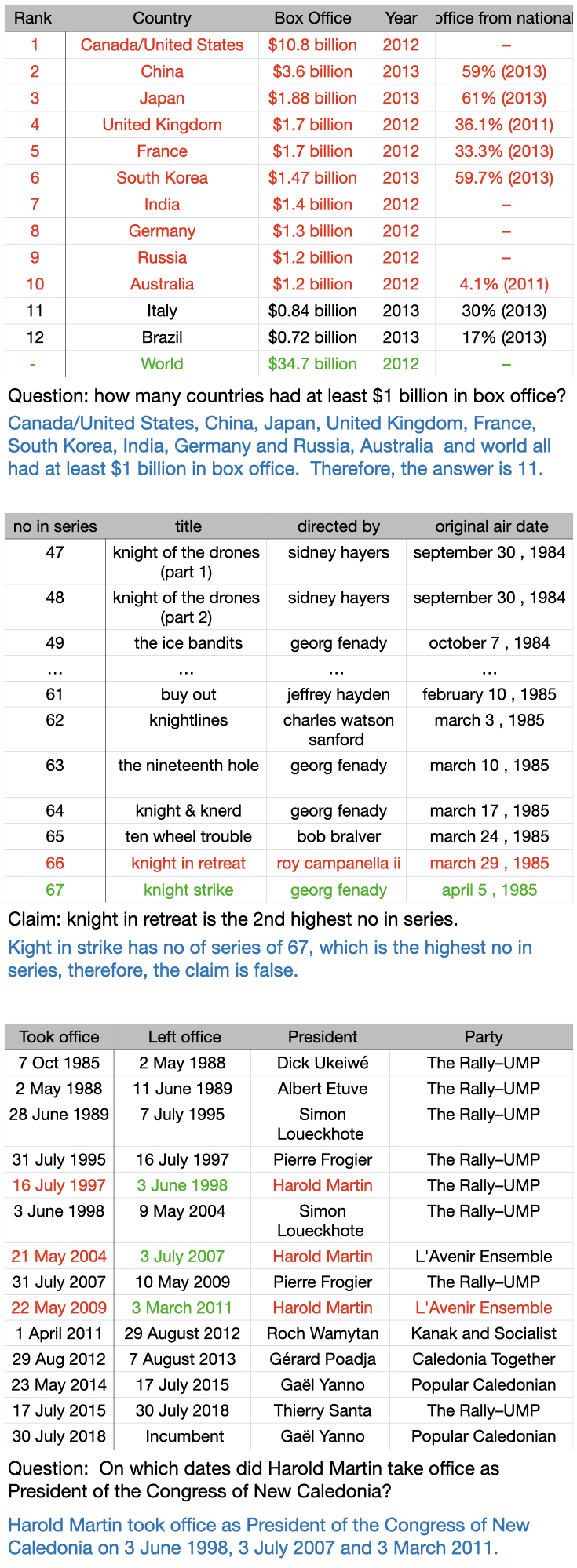}
  \caption{`Wrong' predictions from WikiTableQuestions, TabFact, and FetaQA datasets, where `blue' text are the outputs from the GPT-3, `red' means the region of the correct cell to reference, and `green' means the reference trusted by GPT-3.}
  \label{fig:error-examples}
  \vspace{-2ex}
\end{figure}

We also provide a few mistakes made by GPT-3 in~\autoref{fig:error-examples}. In the first example, GPT-3 miscounts the `number of countries above 1  billion box office' because it misidentifies `world' also as a country.  In the second example, GPT-3 misunderstood `2nd highest' as `highest', which leads to prediction error. In the last example, GPT-3 misunderstands the semantics of the question and answers `left office time' instead of `took office time'. These examples show the typical errors of grounding the inputs to the wrong rows or columns of the table. 

\subsection{Analysis}

\paragraph{Impact of Number of Shots}
First of all, we conduct an ablation study to understand the impact of a number of shots in the final performance. In order to control the budget, we only sample 200 samples from WikiTableQuestions, TabFact and FEVEROUS for this ablation study. As can be seen from~\autoref{fig:ablation}, GPT-3 is not quite sensitive to the number of provided demonstrations. Increasing from 1-shot to 2-shot can often benefit the model, however, increasing the shot number further does not yield more performance gain. We conjecture that instruct fine-tuning used in GPT-3~\cite{ouyang2022training} can easily extrapolate the task meaning, thus, having a single demonstration is already enough for the model to understand the task.  
\begin{figure}[h!]
  \centering
  \begin{tikzpicture}
\begin{axis}[
height=1.2in,
xlabel=num of shots, %横坐标名
ylabel=accuracy, %纵坐标名
tick align=outside, %刻度在外显式
width=0.95\linewidth,
legend style={at={(1,0.5)},anchor=south east,legend columns=-1},
scale only axis,
axis x line*=bottom,
hide y axis,
xticklabel style = {font=\small,yshift=0.5ex},
ymin = 20, 
ymax = 80,
]

\addplot[smooth,mark=triangle,cyan] plot coordinates {
    (0, 20.3)
    (1, 42.2)
    (2, 43.0)
    (3, 41.5)
    (5, 43.2)
};
\addplot[smooth,mark=triangle,red] plot coordinates {
    (0, 62.4)
    (1, 77.3)
    (2, 76.8)
    (3, 77.0)
    (5, 76.2)
};
\addplot[smooth,mark=triangle,blue] plot coordinates {
    (0, 70.4)
    (1, 77.0)
    (2, 76.2)
    (3, 75.3)
    (5, 76.1)
};
\legend{WikiTQ, TabFct, FEV}
\end{axis}
\end{tikzpicture}
  \vspace{-4ex}
  \caption{k-shot ablation study over WikiTableQuestions and TabFact and FEVEROUS.}
  \begin{tikzpicture}
\begin{axis} [
ybar,
height=1.2in, 
bar width=0.4cm,
width=0.95\linewidth,
scale only axis,
ymin = 0, 
yticklabels=\empty,
axis x line*=bottom,
hide y axis,
xticklabel style = {font=\small,yshift=0.5ex},
symbolic x coords={
correct,
wrong,
no reason,
},
legend style={
    at={(0,1.0)},
    anchor=north west,
    legend columns=-1
},
enlarge x limits=0.2,
%title=Breakdown Accuracy over Question Types,
xtick=data,
yticklabels=\empty,
nodes near coords,
nodes near coords align={vertical},
every node near coord/.append style={font=\tiny},
legend style={at={(1,1)},anchor=north east}
]
\addplot coordinates {
(correct, 94) (wrong, 6) (no reason, 0)
};
\addplot coordinates {
(correct, 86) (wrong, 10) (no reason, 4)
};
\addplot coordinates {
(correct, 88) (wrong, 4) (no reason, 8)
};
\legend{WikiTQ, TabFact, FEV}
\end{axis}
\end{tikzpicture}
  \vspace{-4ex}
  \caption{human evaluation of `reasoning chains' in WikiTableQuestions, TabFact, and FEVEROUS.}
  \label{fig:ablation}
  \vspace{-2ex}
\end{figure}

\paragraph{Quality Evaluation of Reasoning Chains}
We conduct a human evaluation to assess whether GPT-3 is making the correct prediction with the correct reasons. Specifically, we sample 100 reasoning paths from the correctly predicted examples and manually study whether these reasoning chains are grounded on the table or simply `hallucination'. As can be seen from~\autoref{fig:ablation}, we found that around 90\% of reasoning chains are faithful to the information in the table, and only less than 10\% of the reasoning chains are hallucinated. Based on this evaluation, we believe that LLMs are not guessing the answers correctly by chance. 

We believe these `reasoning chains' are useful in many aspects: (1) the chains can provide a rationale to humans to justify the decision-making process. (2) one of the notorious annotation tasks is to annotate the `underlying' semantic form for many NLP tasks, which require expertise for human annotators, on the other hand, the annotation cost is huge. Using GPT-3 to demonstrate useful natural language `semantic forms' could potentially greatly lower the annotation burden of these tasks.  

\begin{figure}[h!]
  \centering
  \begin{tikzpicture}
\begin{axis}[
height=1.2in,
xlabel=table size in token length, %横坐标名
ylabel=accuracy, %纵坐标名
tick align=outside, %刻度在外显式
width=0.95\linewidth,
legend style={at={(1,1)},anchor=north east,legend columns=-1},
scale only axis,
axis x line*=bottom,
hide y axis,
xticklabel style = {font=\small,yshift=0.5ex},
ymin = 20, 
ymax = 80,
]
\addplot[smooth,mark=triangle,cyan] plot coordinates {
    (100, 79)
    (200, 77)
    (300, 72)
    (400, 64)
    (500, 66)
    (600, 68)
    (700, 50)
    (800, 50)
    (900, 50)
    (1000, 50)
    (1100, 50)
    (1100, 50)
};
\addplot[smooth,mark=triangle,red] plot coordinates {
    (100, 62)
    (200, 51)
    (300, 48)
    (400, 40)
    (500, 45)
    (600, 50)
    (700, 35)
    (800, 21)
    (900, 1)
    (1000, 1)
    (1100, 1)
    (1100, 1)
};
\legend{TabFct, WikiTQ}
\end{axis}
\end{tikzpicture}
  \vspace{-4ex}
  \caption{Model performance on WikiTableQuestions and TabFact w.r.t the input table size.}
  \label{fig:length}
  \vspace{-2ex}
\end{figure}
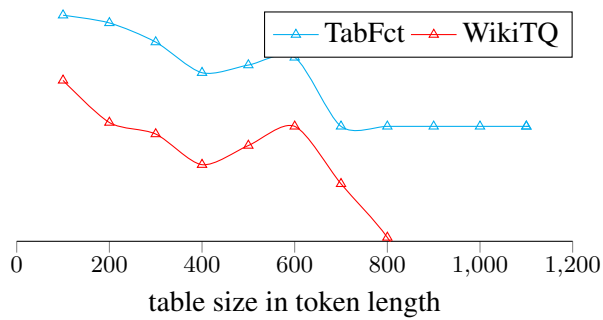

\paragraph{Impact of Table Size}
An important factor for model performance is the size of the table. Here we want to understand how relevant the model performance is w.r.t the input table length. We group the table token length into different groups like `0-100', `100-200', etc, and plot the group-wise accuracy for WikiTables and TabFact in~\autoref{fig:length}. As can be seen from the table, we found that GPT-3's performance is highly sensitive to the table size. As the table size grows, the accuracy almost decreases monotonically. After the table size exceeds 1000 tokens (e.g. 1500 word pieces), GPT-3's performance almost degrades to random guesses. This ablation study reveals one of the drawbacks of using LLMs for table reasoning. To further enhance LLMs' performance, we need to develop better methods to maintain more consistent performance across different-sized tables. 

\paragraph{Discussions}
In this study, we investigate the possibilities of prompting LLMs to perform complex reasoning tasks over tables. However, we do not believe LLM prompting can replace the existing symbolic methods. LLMs have several favorable properties: (1) no annotation is needed, and (2) the functional coverage is broader than symbolic methods. However, LLM prompting exhibits unpredictable randomness and cannot generalize to large tables. In contrast, symbolic models are (1) agnostic to the table size, and (2) can reliably perform designed functions without much randomness. But they in general require a significant amount of annotated data to learn. 

In conclusion, these two types of models are complementary to each other. To push the limit forward, we need to investigate how to combine the merits of these two types of methods. For example, the symbolic methods can perform certain operations to narrow down to a targeted region in the table, and then LLMs can be used to reason over the limited information.

\section{Conclusion}
In this paper, we investigate whether the current LLMs (GPT-3) can be directly utilized to perform table reasoning tasks. Surprisingly, though LLMs are not optimized for table-based tasks, we found these models highly competent in performing complex table reasoning tasks, especially when combined with `chain of thoughts' prompting. We believe this study can open new possibilities for LLM application in table-related tasks to either directly predict the output or to serve as an auxiliary tool for annotating complex intermediate forms. 

\section*{Limitations}
Our approach has several limitations: (1) the proposed approach is still far from state-of-the-art performance, and there is still room for improve before it can be used as an alternative. (2) the method is still costly, we show that the model can only achieve superior performance when scaling up. Smaller-sized models are still weak at table reasoning. Therefore, we need to consider how to empower smaller models with such reasoning capabilities.

\bibliography{custom}
\bibliographystyle{acl_natbib}

\end{document}